\documentclass{article}
\usepackage{spconf,amsmath,graphicx,hyperref}

\usepackage{booktabs}
\usepackage{xcolor}
\usepackage{multirow}

\usepackage{makecell}
\usepackage{pifont}
\newcommand{\cmark}{\ding{51}} 
\newcommand{\xmark}{\ding{55}} 
\usepackage[numbers,sort&compress]{natbib}
\usepackage{csquotes}

\title{Pose Adaptive Dynamic FiLM Modulation for Visual Speech Recognition}
\name{Matthew Kit Khinn Teng, Haibo Zhang, Takeshi Saitoh}
\address{Kyushu Institute of Technology, Japan\\
Kyushu Institute of Technology, Fukuoka 820-8502, Japan\\}
\begin{document}
%
\maketitle
\begin{abstract} 
Head-pose variation introduces substantial appearance transformations in visual speech recognition (VSR), making pose-aware feature modulation desirable. However, performance degradation and unwanted feature interactions may result from using numerous Feature-wise Linear Modulation (FiLM) circuits with fixed modulation intensity. We propose a Pose Adaptive Dynamic FiLM framework with a Dynamic Residual FiLM (DR-FiLM) modulator that predicts input-dependent weights to adaptively control the strength of pose-conditioned modulation. Experiments on LRS2 and LRS3 demonstrate that unweighted multi-pathway modulation substantially degrades phoneme recognition, increasing PER to 20.33\% and 29.42\%, respectively, compared with 16.20\% and 20.96\% for the single ResFiLM configuration. In contrast, the proposed DR-FiLM with dynamic Deep–Res weighting reduces PER to 15.74\% on LRS2 and 23.91\% on LRS3, substantially mitigating the adverse effects of unweighted modulation. The analysis of the learned weights further reveals a consistent tendency to assign greater weight to the deeper FiLM pathway as head-pose variation increases. These results show that merging pose-conditioned FiLM circuits is more efficient when the modulation strength is dynamically controlled.
\end{abstract}
%


\section{Introduction}
\label{sec:intro}
Lip-reading is a form of V-ASR that derives spoken words from visual cues of facial articulatory features, particularly lip movements. Many existing VSR approaches directly predict words or entire sentences from visual speech~\cite{prajwal_2022subword,ma_2023avsr,liu_2023synthvsr,djilali_2023lip2vec,laux_2024litevsr}. However, direct word-level prediction can be challenged by speaker-dependent variations, large vocabularies, and subtle visual differences between words. This has motivated phonetic representations, in which phoneme-level prediction provides a more fine-grained representation of visual speech and reduces word-level ambiguity. Despite this advantage, phoneme-level VSR remains sensitive to appearance variations caused by head-pose changes. Such variations alter the spatial configuration and appearance of the lip region, making recognition more challenging under large or unconstrained viewing angles. Prior studies have addressed this difficulty through multi-view representation learning~\cite{ma_IEEE_2021,isobe_2021mdpi,maeda_2021apsipa,jeon_2022mdpi} and pose-oriented data augmentation or synthetic view generation~\cite{cheng_2020icassp,hao_2025lipgen,fernandez_2023sparsevsr,kang_2023ijcv}; however, these approaches either rely on predefined viewpoints or synthetic transformations, rather than explicitly modeling pose-induced representation changes within the VSR network.

To solve this limitation, our prior research presented HP-VSR-ResFiLM~\cite{teng_2026hpvsr}, which conditions visual feature modulation on estimated head posture. While it generates sample-adaptive modulation parameters ($\alpha$ and $\gamma$) based on head pose, it applies these modulations through a rigid, unweighted structure where all FiLM pathways contribute uniformly regardless of posture variation. This fixed weighting setup may not be optimal for all utterances, as varied head-pose conditions may necessitate different levels of feature modification across layers.This raises a crucial question: can the model dynamically select or weight which pose-conditioned feature pathways are best suited to each input?

Motivated by this observation, our proposed framework extends HP-VSR-ResFiLM~\cite{teng_2026hpvsr} by adaptively weighting multiple pose-conditioned FiLM pathways according to the input. The main contributions of this study are: 1) We introduce a Dynamic Residual FiLM (DR-FiLM) Modulator with a dynamic weighting mechanism that adaptively controls the modulation strength of multiple pose-conditioned FiLM pathways. 2) We study several configurations of Deep FiLM and ResFiLM pathways and show that adaptive weighting can reduce performance loss caused by unweighted multi-pathway modulation. 3) We examine the acquired dynamic weights over several yaw ranges, gaining insight into how the model adjusts its dependence on distinct FiLM routes under varied head-position situations.

\begin{figure}[t] 
  \centering
   \includegraphics[scale=0.162]{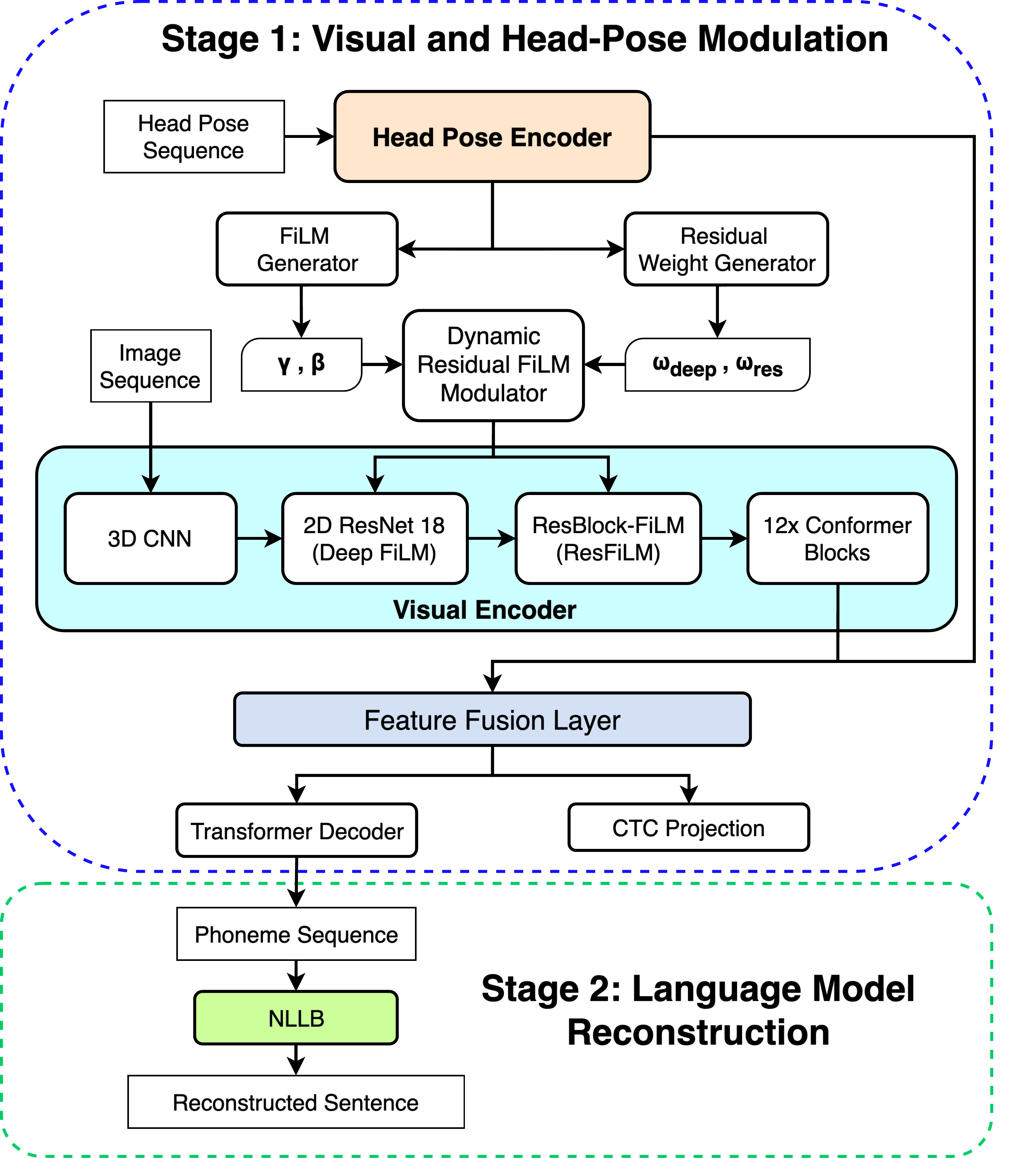} 

   \caption{Overview of the proposed architecture. Head-pose features dynamically determine the contribution of the Deep FiLM pathways at ResNet 18 Layers~3--4 and the ResFiLM pathway for pose-adaptive feature modulation.}
   
   \label{fig:overall_architecture}
\end{figure}

\section{Methodology}
\label{sec:method}
\noindent
\textbf{Framework Overview.}
The proposed framework is shown in Figure~\ref{fig:overall_architecture}. The visual and head-pose encoders retain the same architecture as in \cite{teng_2026hpvsr}. In the visual pathway, the existing FiLM modules at 2D ResNet 18 frontend Layers~3 and~4 (Deep FiLM) are retained, and an additional ResBlock-FiLM (ResFiLM) module is added after the 2D ResNet 18. Moreover, the encoded head-pose features are used by two components: a FiLM generator that produces the modulation parameters $\gamma$ and $\beta$, and a residual weight generator that predicts input-dependent weights $w_{\mathrm{deep}}$ and $w_{\mathrm{res}}$ for the Deep FiLM and ResFiLM pathways, respectively. These parameters and weights are used by the proposed DR-FiLM modulator to adaptively control the strength of pose-conditioned modulation. The resulting visual features are subsequently combined with the head-pose features through an MLP, while the remaining phoneme prediction architecture follows the previous framework.

\noindent
\textbf{Residual Weight Generator.}
The residual weight generator predicts the relative modulation strength of the Deep FiLM and ResFiLM (Deep + Res) pathways from the encoded head-pose feature. It consists of two linear layers with an intermediate ReLU activation. Given the head-pose feature $\mathbf{h}$, the generator produces two unnormalized weights:

\begin{equation}
[\tilde{w}_{\mathrm{deep}},\tilde{w}_{\mathrm{res}}]=f_{\mathrm{w}}(\mathbf{h}),
\end{equation}
which are normalized using a softmax function:
\begin{equation}
[w_{\mathrm{deep}},w_{\mathrm{res}}]=\operatorname{softmax}([\tilde{w}_{\mathrm{deep}},\tilde{w}_{\mathrm{res}}]).
\end{equation}
Thus,
\begin{equation}
w_{\mathrm{deep}}+w_{\mathrm{res}}=1.
\end{equation}
These input-dependent weights are provided to the DR-FiLM Modulator to adaptively control the strength of pose-conditioned modulation in each pathway.

\noindent
\textbf{DR-FiLM Modulator.}
This modulator combines pose-conditioned feature modulation with input-dependent modulation strength. Two pose-conditioned modulation pathways are employed: a Deep FiLM pathway applied at ResNet 18 Layers~3 and~4, and an additional ResFiLM pathway. 

For the 2D ResNet 18 (Deep FiLM) pathway, the proposed DR-FiLM operation is defined as
\begin{equation}
\mathbf{F}_{\mathrm{DR-FiLM}}^{\mathrm{deep}}=\mathbf{F}_{\mathrm{out}}^{\mathrm{deep}}+w_{\mathrm{deep}}
\left(\mathbf{F}_{\mathrm{out}}^{\mathrm{deep}}\odot\gamma_{\mathrm{deep}}+\beta_{\mathrm{deep}}\right),
\end{equation}
where $\mathbf{F}_{\mathrm{out}}^{\mathrm{deep}}$ denotes the output feature of the corresponding convolutional transformation. This operation is applied sequentially at 2D ResNet 18 Layers~3 and~4.

Similarly, the additional ResBlock-FiLM (ResFiLM) pathway is defined as
\begin{equation}
\mathbf{F}_{\mathrm{DR-FiLM}}^{\mathrm{res}}=\mathbf{F}_{\mathrm{out}}^{\mathrm{res}}+w_{\mathrm{res}}\left(\mathbf{F}_{\mathrm{out}}^{\mathrm{res}}\odot\gamma_{\mathrm{res}}+\beta_{\mathrm{res}}\right).
\end{equation}

Unlike conventional FiLM~\cite{perez_2018film} and directly gated FiLM formulations~\cite{lin_2021task}, which were originally developed for different applications, the proposed DR-FiLM adopts their feature modulation formulation while introducing a dynamically weighted pose-conditioned modulation term for VSR. Specifically, DR-FiLM retains the original feature representation as a base term and introduces dynamically weighted modulation through the Deep FiLM and ResFiLM pathways. The dynamic weights $w_{\mathrm{deep}}$ and $w_{\mathrm{res}}$ therefore control the strength of pose-conditioned modulation in the respective pathways.

\begin{table*}[t]
\centering
\caption{Performance comparison (PER \% and WER \%) of representative VSR methods and our proposed framework on the LRS2 and LRS3 datasets. The total hours indicate the combined duration or size of the pretraining and training datasets.}

\label{tab:sota_comparison}
\setlength{\tabcolsep}{6pt}
\begin{tabular}{l c c c c c c }
\toprule
\multirow{2}{*}{Method} & \multirow{2}{*}{Modality} & \multirow{2}{*}{FiLM Layers} & \multicolumn{2}{c}{LRS2} & \multicolumn{2}{c}{LRS3} \\
\cmidrule(lr){4-5} \cmidrule(lr){6-7}
& & & Total Hours & WER $\downarrow$ & Total Hours & WER $\downarrow$\\
\midrule
Hyb.-Conf. \cite{ma_IEEE_2021} & Video & -- & 223 & 39.1 & 438 & 46.9 \\
Hyb.-Conf. \cite{ma_IEEE_2021} & Video & -- & 381 & 37.9 & 590 & 43.3 \\
VTP \cite{prajwal_2022subword} & Video & --  & 2676 & 22.6 & 2676 & 30.7 \\
Auto-AVSR \cite{ma_2023avsr} & Video & -- & 818 & 27.9 & 818 & 33.0 \\
Auto-AVSR \cite{ma_2023avsr} & Video & -- & 3448 & 14.6 & 3448 & 19.1 \\
CM-aux \cite{ma_2022multilanguage} & Video & -- & 223 & 32.9 & 438 & 37.9 \\
SyncVSR \cite{ahn_2024syncvsr}  & Video$^\dagger$ & -- & 223 & 28.9 & 438 & 31.2 \\
GLip \cite{wang_2025glip} & Video  & --   & 223 & 27.4 & 438 & 30.1 \\
PV-ASR~\cite{teng_2026hpvsr} & Video + 117-Point  & --   & 223 & 26.6 & 438 & 36.7 \\
HP-VSR-ResFiLM \cite{teng_2026hpvsr} & Video + Headpose & ResFiLM  & 223 & 24.7  & 438 & 30.3 \\
HP-VSR-FiLMFuse \cite{teng_2026hpvsr} & Video + Headpose & L3--L4  & 223 & 25.4  & 438 & 32.5 \\
Ours (Baseline) & Video + Headpose & L3--L4 + ResFiLM   & 223 & 28.8 & 438 & 39.0 \\
Ours (Abs FiLM (R$\rightarrow$D)) & Video + Headpose & L3--L4 + ResFiLM   & 223 & 27.6 & 438 & 33.9 \\
Ours (DR-FiLM (Deep + Res)) & Video + Headpose & L3--L4 + ResFiLM   & 223 & 23.2 & 438 & 34.4 \\ 
\bottomrule
\end{tabular}
\vspace{2pt}
{\footnotesize $^\dagger$Audio used only as an auxiliary training signal (crossmodal token prediction); inference is video-only.}
\end{table*}

\noindent
\textbf{Absolute-Pose FiLM Baseline.}
To investigate whether the performance of dynamic weighting can be attributed directly to head-pose magnitude, we additionally construct an absolute-pose-based FiLM baseline (Abs FiLM). Unlike DR-FiLM, the weights in Abs FiLM are not predicted by the residual weight generator. Instead, they are assigned deterministically based on the average absolute yaw angle across all frames. Specifically, let $p = \frac{1}{T} \sum \vert{}\text{yaw}_t\vert{}$ denote the mean absolute yaw over $T$ frames, and define
\begin{equation}
\alpha =
\begin{cases}
0, & p=0^\circ,\\
p/30^\circ, & 0^\circ < p < 30^\circ,\\
1, & p\geq30^\circ.
\end{cases}
\end{equation}
Two complementary weighting directions are considered. For Abs FiLM (R$\rightarrow$D), the weights are defined as $w_{\mathrm{res}}=1-\alpha$ and $w_{\mathrm{deep}}=\alpha$, whereas for Abs FiLM (D$\rightarrow$R), they are reversed as $w_{\mathrm{res}}=\alpha$ and $w_{\mathrm{deep}}=1-\alpha$. Thus, the two variants respectively shift the weighting from ResFiLM to Deep FiLM and from Deep FiLM to ResFiLM as the absolute yaw magnitude increases, while maintaining $w_{\mathrm{deep}}+w_{\mathrm{res}}=1$.

\section{Experimental Setup}
\label{sec:experiment}

\noindent
\textbf{Datasets.}
Experiments are conducted on the publicly available English visual speech datasets LRS2~\cite{son_2017lrs2}, LRS3~\cite{afouras_2018lrs3}. LRS2 contains approximately 144k utterances (224.5 h), while LRS3 comprises approximately 152k utterances (438.9 h), with both datasets providing separate pretraining, training/validation, and test partitions.

\noindent
\textbf{Experimental Configuration.}
The training and evaluation protocols follow our previous HP-VSR framework~\cite{teng_2026hpvsr}, including the Stage~1 initialization, optimization settings, batch configuration, and checkpoint averaging. Stage~2 uses the same fixed pretrained NLLB model as in~\cite{teng_2026pvsr} without additional training or fine-tuning.

\noindent
\textbf{Evaluation Metrics.} 
VSR performance is evaluated using Word Error Rate (WER)~\cite{jelinek_1975wer}, which measures the proportion of word-level deletions, insertions, and substitutions relative to the reference text. Phoneme Error Rate (PER) is reported to assess recognition performance at the phoneme level.

\section{Experimental Results}
\label{sec:results}

\noindent
\textbf{Comparison of Recent Representative Methods.}
Table~\ref{tab:sota_comparison} shows that our DR-FiLM (Deep + Res) configuration achieves the strongest LRS2 result among all video-only methods at comparable data scale (23.2\% WER), outperforming GLip~\cite{wang_2025glip}, CM-aux~\cite{ma_2022multilanguage}, and our own previous work, HP-VSR-ResFiLM~\cite{teng_2026hpvsr}, while substantially recovering the degradation observed in the unweighted multi-pathway configuration (28.8\% $\rightarrow$ 23.2\% on LRS2, 39.0\% $\rightarrow$ 34.4\% on LRS3). This confirms that dynamic, input-dependent weighting is essential when combining multiple pose-conditioned FiLM pathways, consistent with our ablation findings in Table~\ref{tab:dynamic_film_ablation}. However, on LRS3, DR-FiLM (34.4\%) does not surpass GLip (30.1\%) or our prior single-location ResFiLM variant (30.3\%), despite its clear advantage on LRS2. We attribute this discrepancy to differences in recording conditions between the two benchmarks: LRS2's BBC broadcast footage and LRS3's TED-talk footage differ in pose distribution, lighting, and speaker framing, which may affect how the dual-pathway modulation interacts with each domain. We view improving cross-dataset consistency of DR-FiLM as an important direction for future work.

\begin{table}[t]
\centering
\caption{Ablation study of DR-FiLM configurations and dynamic weighting on LRS2 and LRS3. D denotes the Deep FiLM pathway applied at ResNet 18 Layers~3--4, while R denotes the ResFiLM pathway applied after the 2D CNN frontend. R$\rightarrow$D shifts the weighting from ResFiLM toward Deep FiLM as the absolute yaw angle increases, whereas D$\rightarrow$R applies the reverse weighting. }

\label{tab:dynamic_film_ablation}
\resizebox{\linewidth}{!}{

\begin{tabular}{l c c c c c }
\toprule
\multirow{2}{*}{Method} & \multirow{2}{*}{Dynamic } & \multicolumn{2}{c}{LRS2} & \multicolumn{2}{c}{LRS3} \\
\cmidrule(lr){3-4} \cmidrule(lr){5-6}
& Weights & PER $\downarrow$ & WER $\downarrow$ & PER $\downarrow$ & WER $\downarrow$\\
\midrule
Baseline & \xmark  & 20.33 & 28.84 & 29.42 & 38.98 \\
DR-FiLM (Layer-wise) & \cmark  & 20.14 & 29.01 & 25.88 & 35.54 \\
DR-FiLM (Deep + Res) & \cmark  & 15.74 & 23.24 & 23.91 & 34.43 \\ 
DR-FiLM (L4 + Res) & \cmark  & 16.79 & 24.98 & 27.08 & 37.58 \\
Abs FiLM (R$\rightarrow$D) & \cmark  & 18.79 & 27.55 & 23.67 & 33.90 \\
Abs FiLM (D$\rightarrow$R) & \cmark  & 19.76 & 28.44 & 23.77 & 33.82 \\	
\bottomrule
\end{tabular}
}
\end{table}

\textbf{Analysis of DR-FiLM Configurations.}Table~\ref{tab:dynamic_film_ablation} evaluates various dynamic-weighting configurations—including layer-wise weighting ($w_3+w_4+w_{\mathrm{res}}=1$) and L4+Res weighting ($w_4+w_{\mathrm{res}}=1$)—to analyze the impact of weighting granularity and modulation depth. Simply combining multiple FiLM pathways without adaptive weighting degrades performance significantly. For instance, the unweighted L3--L4 + ResFiLM baseline increases error rates on LRS2 (PER/WER rising from 16.20\%/24.68\% to 20.33\%/28.84\%) and LRS3 (from 20.96\%/30.32\% to 29.42\%/38.98\% compared to single HP-VSR-ResFiLM). This indicates that concurrently activating multiple pose-conditioned routes without coordination leads to feature interference. Dynamic weighting effectively mitigates this degradation, though performance varies by setup. While the layer-wise configuration yields mixed results (20.14\%/29.01\% on LRS2; 25.88\%/35.54\% on LRS3), the proposed DR-FiLM (Deep + Res) configuration achieves the best overall performance among dynamic variants (15.74\%/23.24\% on LRS2; 23.91\%/34.43\% on LRS3). Comparing this with the DR-FiLM (L4+Res) setup (16.79\%/24.98\% and 27.08\%/37.58\%) confirms that jointly weighting the deeper L3–L4 layers is superior to restricting adaptation to L4 alone. Finally, the Abs-FiLM variants test deterministic weighting based on absolute yaw. While both directional rules ($R\rightarrow{}D$: 18.79\%/27.55\% and 23.67\%/33.90\%; $D\rightarrow{}R$: 19.76\%/28.44\% and 23.77\%/33.82\%) outperform the unweighted baseline, they remain inferior to learned DR-FiLM (Deep + Res). This demonstrates that input-dependent learned weighting is more robust than a rigid, heuristic pose-rule. Overall, DR-FiLM (Deep + Res) successfully recovers the degradation caused by unweighted multi-pathway modulation.

\begin{table}[t]
\centering
\caption{Average DR-FiLM (Deep + Res) weights and PER across different head-pose ranges on LRS2 and LRS3. $w_{\mathrm{deep}}$ and $w_{\mathrm{res}}$ denote the weights assigned to the deeper L3--L4 FiLM and ResFiLM branches, respectively. The standard deviation (Std.) is identical for both weights because $w_{\mathrm{deep}}+w_{\mathrm{res}}=1$.}
\label{tab:dynamic_film_weights}
  \resizebox{\linewidth}{!}{
\begin{tabular}{l c c c c c c}
\toprule
Dataset & \shortstack{Yaw\\Pose}  & \shortstack{Avg. \\$w_{\mathrm{deep}}$} & \shortstack{Avg. \\$w_{\mathrm{res}}$} & Std. & \shortstack{Avg. \\PER} & \shortstack{No. of\\Samples} \\
\midrule
\midrule
LRS2 & $<15^\circ$ & 0.7948 & 0.2052 & 0.0043 & 14.03 & 693 \\
& $15$--$30^\circ$ & 0.8027 & 0.1973 & 0.0068 & 17.50 & 341 \\
& $\geq 30^\circ$  & 0.8234 & 0.1766 & 0.0162 & 19.36 & 209 \\
\midrule
LRS3 & $<15^\circ$ & 0.7586 & 0.2414 & 0.0031 & 23.22 & 430 \\
& $15$--$30^\circ$ & 0.7631 & 0.2369 & 0.0050 & 23.35 & 488 \\
& $\geq 30^\circ$  & 0.7794 & 0.2206 & 0.0168 & 25.42 & 403 \\
\bottomrule
\end{tabular}
}
\end{table}

\noindent
\textbf{Analysis of Dynamic Weights.}
Table~\ref{tab:dynamic_film_weights} presents the average learned dynamic weights across different head-pose ranges for the LRS2 and LRS3 datasets. On LRS2, the average $w_{\mathrm{deep}}$ increases from 0.7948 for samples with $|yaw|<15^\circ$ to 0.8027 for $15^\circ\leq|yaw|<30^\circ$ and 0.8234 for $|yaw|\geq30^\circ$. A similar trend is observed on LRS3, where $w_{\mathrm{deep}}$ increases from 0.7586 to 0.7631 and 0.7794 across the same pose ranges. Correspondingly, $w_{\mathrm{res}}$ decreases as the pose range increases, since the two weights are constrained by $w_{\mathrm{deep}}+w_{\mathrm{res}}=1$.

These findings show that the model assigns greater modulation strength to the deeper FiLM route under larger head-pose fluctuations, while decreasing the relative strength of the ResFiLM pathway. This behavior indicates that the suggested dynamic weighting method learns to alter modulation strength based on the input pose, rather than applying a set weighting to all samples. The increase in $w_{\mathrm{deep}}$ is more noticeable for highly posed samples, notably on LRS2. This suggests that deeper feature modulation may become more relevant as pose variation grows.

\section{Limitations and Future Work}

Although our Dynamic FiLM framework mitigates unweighted multi-pathway degradation, it does not uniformly outperform the strongest HP-VSR-ResFiLM baseline across both benchmarks: while DR-FiLM (Deep + Res) surpasses the baseline on LRS2, a performance gap persists on LRS3. This indicates that dynamic weighting efficacy is sensitive to dataset characteristics like pose distribution and video quality.

Future work will explore more expressive weighting mechanisms, extend utterance-level routing to frame-level temporal adaptation for within-utterance pose changes, and incorporate additional geometry (pitch and roll). Finally, broader cross-domain evaluations will help verify generalization across diverse recording conditions.

\section{Conclusions}

This study proposes a Pose-Adaptive Dynamic FiLM framework with DR-FiLM to dynamically control multiple pose-conditioned FiLM pathways for VSR. Experiments on LRS2 and LRS3 show that unweighted multi-pathway modulation can substantially degrade recognition performance, while dynamic weighting effectively mitigates this degradation. The proposed DR-FiLM (Deep + Res) configuration achieves PER/WERs of 15.74\%/23.24\% on LRS2 and 23.91\%/34.43\% on LRS3, demonstrating the effectiveness of dynamically balancing the contributions of Deep FiLM and ResFiLM. The analysis reveals that the pathway weights vary with head-pose severity, implying that DR-FiLM can adapt feature modulation according to pose conditions rather than relying on a fixed configuration. These results highlight dynamic pose-conditioned weighting as an effective approach for improving the performance of phoneme-based VSR under varying head poses.

\section*{Acknowledgments}

This work was supported by JSPS KAKENHI Grant Number JP23H03787.

\vfill\pagebreak

\bibliographystyle{IEEEtran}
\bibliography{paper}

\end{document}